# Parallel transport in shape analysis: a scalable numerical scheme

Maxime Louis[†12], Alexandre Bône[†12], Benjamin Charlier[23], Stanley Durrleman[12], and the Alzheimer's Disease Neuroimaging Initiative

[1] Sorbonne Universités, UPMC Université Paris 06, Inserm, CNRS, Institut du Cerveau et de la Moelle (ICM) – Hôpital Pitié-Salpêtrière, 75013 Paris, France,
[2] Inria Paris, Aramis project-team, 75013 Paris, France,
[3] Université de Montpellier, France

**Abstract.** The analysis of manifold-valued data requires efficient tools from Riemannian geometry to cope with the computational complexity at stake. This complexity arises from the always-increasing dimension of the data, and the absence of closed-form expressions to basic operations such as the Riemannian logarithm. In this paper, we adapt a generic numerical scheme recently introduced for computing parallel transport along geodesics in a Riemannian manifold to finite-dimensional manifolds of diffeomorphisms. We provide a qualitative and quantitative analysis of its behavior on high-dimensional manifolds, and investigate an application with the prediction of brain structures progression.

## 1 Introduction

Riemannian geometry is increasingly meeting applications in statistical learning. Indeed, working in flat space amounts to neglecting the underlying geometry of the laws which have produced the considered data. In other words, such a simplifying assumption ignores the intrinsic constraints on the observations. When prior knowledge is available, top-down methods can express invariance properties as group actions or smooth constraints and model the data as points in quotient spaces, as for Kendall shape space. In other situations, manifold learning can be used to find a low-dimensional hypersurface best describing a set of observations.

Once the geometry has been modeled, classical statistical approaches for constrained inference or prediction must be adapted to deal with structured data, as it is done in [4,5,11,13]. Being an isometry, the parallel transport arises as a natural tool to compare features defined at different tangent spaces.

In a system of coordinates, the parallel transport is defined as the solution to an ordinary differential equation. The integration of this equation requires to compute the Christoffel symbols, which are in general hard to compute –e.g. in the case of the Levi-Civita connection– and whose number is cubic in the dimension. The Schild's ladder [5], later improved into the Pole ladder [7] when transporting along geodesics, is a more geometrical approach which only requires the computation of Riemannian exponentials and logarithms. When the geodesic equation is autonomous, the scaling and squaring procedure [6] allows to compute exponentials very efficiently. In Lie groups, the Baker-Campbell Haussdorff formula allows fast computations of logarithms with a controlled precision. In such settings, the Schild's ladder is computationnally tractable. However, no theoretical study has studied the numerical approximations or has provided a convergence result. In addition, in the more general case of Riemannian manifolds, the needed logarithm operators are often computationally intractable.

---

[†]Equal contributions.

The Large Deformation Diffeomorphic Metric Mapping (LDDMM) framework [1] focuses on groups of diffeomorphisms, for shape analysis. Geodesic trajectories can be computed by integrating the Hamiltonian equations, which makes the exponential operator computationally tractable, when the logarithm remains costly and hard to control in its accuracy. In [12] is suggested a numerical scheme which approximates the parallel transport along geodesics using only the Riemannian exponential and the metric. The convergence is proved in [8].

In this paper, we translate this so-called *fanning sheme* to finite-dimensional manifolds of diffeomorphisms built within the LDDMM framework [2]. We provide a qualitative and quantitative analysis of its behavior, and investigate a high-dimensional application with the prediction of brain structures progression. Section 2 gives the theoretical background and the detailed steps of the algorithm, in the LDDMM context. Section 3 describes the considered application and discusses the obtained results. Section 4 concludes.

## 2 Theoretical background and practical description

### 2.1 Notations and assumptions

Let $\mathcal{M}$ be a finite-dimensional Riemannian manifold with metric $g$ and tangent space norm $\|\cdot\|_g$. Let $\gamma : t \to [0,1]$ be a geodesic whose coordinates are known at all time. Given $t_0, t \in [0,1]$, the parallel transport of a vector $w \in T_{\gamma(s)}\mathcal{M}$ from $\gamma(t_0)$ to $\gamma(t)$ along $\gamma$ will be noted $\mathrm{P}_{t_0,t}(w) \in T_{\gamma(t)}\mathcal{M}$. We recall that this mapping is uniquely defined by the integration from $u = t_0$ to $t$ of the differential equation $\nabla_{\dot{\gamma}(u)}\mathrm{P}_{t_0,u}(w) = 0$ with $\mathrm{P}_{t_0,t_0}(w) = w$ where $\nabla$ is the Levi-Civita covariant derivative.

We denote Exp the exponential map, and for $h$ small enough we define $\mathrm{J}_{\gamma(t)}^w(h)$, the Jacobi Field emerging from $\gamma(t)$ in the direction $w \in T_{\gamma(t)}\mathcal{M}$ by:

$$\mathrm{J}_{\gamma(t)}^w(h) = \left.\frac{\partial}{\partial \varepsilon}\right|_{\varepsilon=0} \mathrm{Exp}_{\gamma(t)}\big(h\left[\dot{\gamma}(t) + \varepsilon w\right]\big) \in T_{\gamma(t+h)}\mathcal{M}. \qquad (1)$$

### 2.2 The key identity

The following proposition relates the parallel transport to a Jacobi field [12]:

**Proposition.** *For all $t > 0$ and $w \in T_{\gamma(0)}\mathcal{M}$, we have:*

$$\mathrm{P}_{0,t}(w) = \frac{\mathrm{J}_{\gamma(0)}^w(t)}{t} + \mathrm{O}\big(t^2\big). \qquad (2)$$

*Proof.* Let $X(t)$ be the time-varying vector field corresponding to the parallel transport of $w$, i.e. such that $\dot{X}^i + \Gamma^i_{kl} X^l \dot{\gamma}^k = 0$ with $X(0) = w$. At $t=0$, in normal coordinates the differential equation simplifies into $\dot{X}^i(0) = 0$. Besides, near $t=0$ in the same local chart, the Taylor expansion of $X(t)$ writes $X^i(t) = w^i + \mathrm{O}(t^2)$. Noticing that the $i$th normal coordinate of $\mathrm{Exp}_{\gamma(0)}\left(t\left[\dot{\gamma}(t) + \varepsilon w\right]\right)$ is $t(v_0^i + \varepsilon w^i)$, the $i$th coordinate of $\mathrm{J}^w_{\gamma(0)}(t) = \frac{\partial}{\partial \varepsilon}|_{\varepsilon=0} \mathrm{Exp}_{\gamma(0)}\left(t\left[\dot{\gamma}(0) + \varepsilon w\right]\right)$ is therefore $tw^i$, and we thus obtain the desired result. □

Subdividing $[0,1]$ into $N$ intervals and iteratively computing the Jacobi fields $\frac{1}{N} \mathrm{J}^w_{\gamma(k/N)}(\frac{1}{N})$ should therefore approach the parallel transport $\mathrm{P}_{0,1}(w)$. With an error in $\mathrm{O}(\frac{1}{N^2})$ at each step, a global error in $\mathrm{O}(\frac{1}{N})$ can be expected. We propose below an implementation of this scheme in the context of a manifold of diffeomorphisms parametrized by control points and momenta. Its convergence with a rate of $\mathrm{O}(\frac{1}{N})$ is proved in [8].

### 2.3 The chosen manifold of diffeomorphisms

The LDDMM-derived construction proposed in [2] provides an effective way to build a finite-dimensional manifold of diffeomorphisms acting on the $d$-dimensional ambient space $\mathbb{R}^d$. Time-varying vector fields $v_t(.)$ are generated by the convolution of a Gaussian kernel $k(x,y) = \exp\left[-\|x-y\|^2/2\sigma^2\right]$ over $n_{cp}$ time-varying control points $c(t) = [c_i(t)]_i$, weighted by $n_{cp}$ associated momenta $\alpha(t) = [\alpha_i(t)]_i$, i.e. $v_t(.) = \sum_{i=1}^{n_{cp}} k\left[., c_i(t)\right] \alpha_i(t)$. The set of such vector fields forms a Reproducible Kernel Hilbert Space (RKHS).

Those vector fields are then integrated along $\partial_t \phi_t(.) = v_t[\phi(.)]$ from $\phi_0 = \mathrm{Id}$ into a flow of diffeomorphisms. In [10], the authors showed that the kernel-induced distance between $\phi_0$ and $\phi_1$ –which can be seen as the deformation kinetic energy– is minimal i.e. the obtained flow is geodesic when the control points and momenta satisfy the Hamiltonian equations :

$$\dot{c}(t) = \mathrm{K}_{c(t)} \alpha(t), \quad \dot{\alpha}(t) = -\frac{1}{2} \mathrm{grad}_{c(t)} \left\{ \alpha(t)^T \mathrm{K}_{c(t)} \alpha(t) \right\}, \tag{3}$$

where $K_{c(t)}$ is the kernel matrix. A diffeomorphism is therefore fully parametrized by its initial control points $c$ and momenta $\alpha$.

Those Hamiltonian equations can be integrated with a Runge-Kutta scheme without computing the Christoffel symbols, thus avoiding the associated curse of dimensionality. The obtained diffeomorphisms then act on shapes embedded in $\mathbb{R}^d$, such as images or meshes.

For any set of control points $c = (c_i)_{i \in \{1,..,n\}}$, we define the finite-dimensional subspace $V_c = \mathrm{span}\left\{k(.,c_i)\xi \,|\, \xi \in \mathbb{R}^d,\ i \in \{1,..,n\}\right\}$ of the vector fields' RKHS. We fix an initial set $c = (c_i)_{i \in \{1,..,n\}}$ of distinct control points and define the set $\mathcal{G}_c = \{\phi_1 \,|\, \partial_t \phi_t = v_t \circ \phi_t,\ v_0 \in V_c,\ \phi_0 = \mathrm{Id}\}$. Equipped with $\mathrm{K}_{c(t)}$ as –inverse– metric, $\mathcal{G}_c$ is a Riemannian manifold such that $T_{\phi_1} \mathcal{G}_c = V_{c(1)}$, where for all $t$ in $[0,1]$, $c(t)$ is obtained from $c(0) = c$ through the Hamiltonian equations (3) [9].

## 2.4 Summary of the algorithm

We are now ready to describe the algorithm on the Riemannian manifold $\mathcal{G}_c$.

**Algorithm.** Divide $[0,1]$ into $N$ intervals of length $h = \frac{1}{N}$ where $N \in \mathbb{N}$. We note $\omega_k$ the momenta of the transported diffeomorphism, $c_k$ the control points and $\alpha_k$ the momenta of the geodesic $\gamma$ at time $\frac{k}{N}$. Iteratively :

(i) Compute the main geodesic control points $c_{k+1}$ and momenta $\alpha_{k+1}$, using a Runge-Kutta 2 method.
(ii) Compute the control points $c_{k+1}^{\pm h}$ of the perturbed geodesics $\gamma_{\pm h}$ with initial momenta and control points $(\alpha_k \pm h\omega_k, c_k)$, using a Runge-Kutta 2 method.
(iii) Approximate the Jacobi field $\mathrm{J}_{k+1}$ by central finite difference :

$$\mathrm{J}_{k+1} = \frac{c_{k+1}^{+h} - c_{k+1}^{-h}}{2h}. \qquad (4)$$

(iv) Compute the transported momenta $\tilde{\omega}_{k+1}$ according to equation (2) :

$$\mathrm{K}_{c_{k+1}} \tilde{\omega}_{k+1} = \frac{\mathrm{J}_{k+1}}{h}. \qquad (5)$$

(v) Correct this value with $\omega_{k+1} = \beta_{k+1}\tilde{\omega}_{k+1} + \delta_{k+1}\alpha_{k+1}$, where $\beta_{k+1}$ and $\delta_{k+1}$ are normalization factors ensuring the conservation of $\|\omega\|_{V_c} = \omega_k^T K_{c_k} \omega_k$ and of $\langle \alpha_k, \omega_k \rangle_{c_k} = \alpha_k^T K_{c_k} \omega_k$.

As step of the scheme is illustrated in Figure 1. The Jacobi field is computed with only four calls to the Hamiltonian equations. This operation scales quadratically with the dimension of the manifold, which makes this algorithm practical in high dimension, unlike Christoffel-symbol-based solutions. Step (iv) –solving a linear system of size $n_{cp}$– is the most expensive one, but remained within reasonable computational time in the investigated examples.

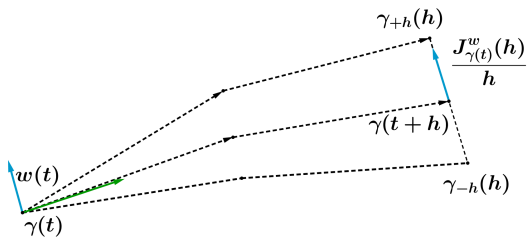

Fig. 1: Step of the parallel transport of the vector $w$ (blue arrow) along the geodesic $\gamma$ (solid black curve). $\mathrm{J}_\gamma^w$ is computed by central finite difference with the perturbed geodesics $\gamma_\varepsilon$ and $\gamma_-$, integrated with a second-order Runge-Kutta scheme (dotted black arrows). A fan of geodesics is formed.

In [8], the authors prove the convergence of this scheme, and show that the error increases linearly with the size of the step used. The convergence is guaranteed as long as the step (ii) is performed with a method of order at least two. A first order method in step (iii) is also theoretically sufficient to guarantee convergence. Those variations will be studied in Section 3.3.

## 3  Application to the prediction of brain structures

### 3.1  Introducing the exp-parallelization concept

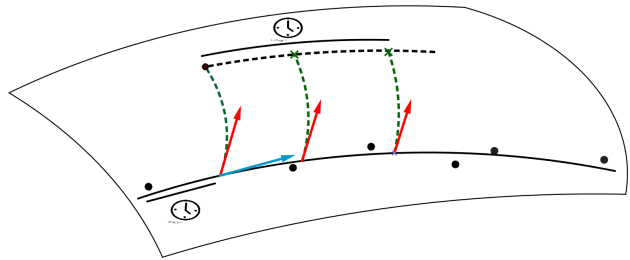

Fig. 2: Time-reparametrized *exp-parallelization* of a reference geodesic model. The black dots are the observations, on which are fitted a geodesic regression (solid black curve, parametrized by the blue arrow) and a matching (leftmost red arrow). The red arrow is then parallel-transported along the geodesic, and exponentiated to define the *exp-parallel* curve (black dashes).

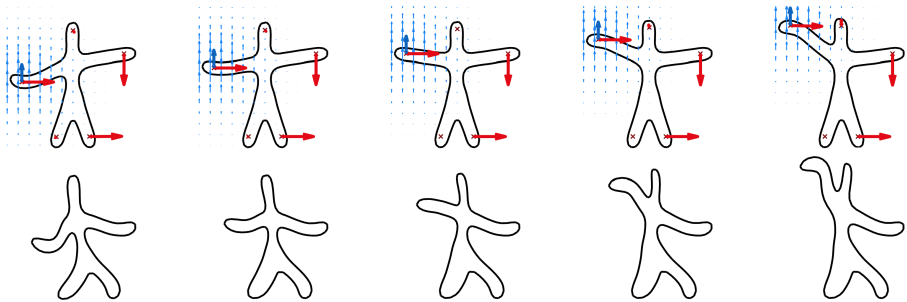

Fig. 3: Illustration of the *exp-parallelization* concept. Top row: the reference geodesic at successive times. Bottow row: the exp-parallel curve. Blue arrows: the geodesic momenta and velocity field. Red arrows: the momenta describing the initial registration with the target shape and its transport along the geodesic.

Exploiting the fanning scheme described in Section 2.4, we can parallel-transport any set of momenta along any given reference geodesic. Figure 2 illustrates the procedure. The target shape is first registered to the reference geodesic : the diffeomorphism that best transforms the chosen reference shape into the target one is estimated with a gradient descent algorithm on the initial control points and momenta [2]. Such a procedure can be applied generically to images or meshes. Once this geodesic is obtained, its initial set of momenta is parallel-transported along the reference geodesic. Taking the Riemannian exponential of the transported vector at each point of the geodesic defines a new trajectory, which we will call *exp-parallel* to the reference one.

As pointed out in [5], the parallel transport is quite intuitive in the context of shape analysis, for it is an isometry which transposes the evolution of a shape into the geometry of another shape, as illustrated by Figure 3.

### 3.2 Data and experimental protocol

Repeated Magnetic Resonance Imaging (MRI) measurements from 71 subjects are extracted from the ADNI database and preprocessed through standard pipelines into affinely co-registered surface meshes of hippocampi, caudates and putamina. The geometries of those brain sub-cortical structures are altered along the Alzheimer's disease course, which all considered subjects finally convert to.

Two subjects are successively chosen as references, for they have fully developed the disease within the clinical measurement protocol. As illustrated on Figure 2, a geodesic regression [3] is first performed on each reference subject to model the observed shape progression. The obtained trajectory on the chosen manifold of diffeomorphisms is then *exp-parallelized* into a shifted curve, which is hoped to model the progression of the target subject.

To account for the variability of the disease dynamics, for each subject two scalar coefficients encoding respectively for the disease onset age and the rate of progression are extracted from longitudinal cognitive evaluations as in [11]. The exp-parallel curve is time-reparametrized accordingly, and finally gives predictions for the brain structures. In the proposed experiment, the registrations and geodesic regressions typically feature around 3000 control points in $\mathbb{R}^3$, so that the deformation can be seen as an element of a manifold of dimension 9000.

### 3.3 Estimating the error associated to a single parallel transport

To study the error in this high-dimensional setting, we compute the parallel transport for a varying number of discretization steps $N$, thus obtaining increasingly accurate estimations. We then compute the empirical relative errors, taking the most accurate computation as reference.

Arbitrary reference and target subjects being chosen, Figure 4 gives the results for the proposed algorithm and three variations : without enforcing the conservations at step (v) [WEC], using a Runge-Kutta of order 4 at step (ii) [RK4], and using a single perturbed geodesic to compute $J$ at step (iii) [SPG].

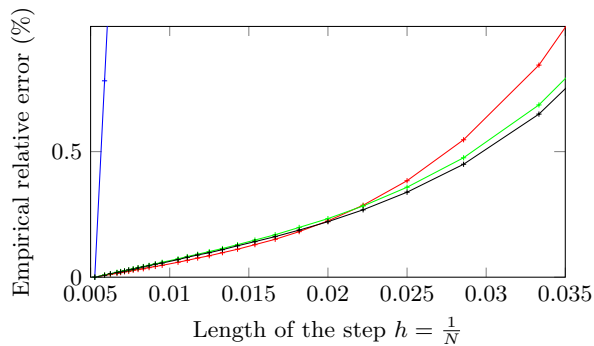

Fig. 4: Empirical relative error of the parallel transport in a high-dimensional setting.
In black the proposed algorithm, in green the WEC variant, in red the RK4 variant, and in blue the SPG one.

We recover a linear behavior with the length of the step $\frac{1}{N}$ in all cases. The SPG variant converges much slower, and is excluded from the following considerations.

For the other algorithms, the empirical relative error remains below 5% with 15 steps or more, and below 1% with 25 steps or more. The slopes of the asymptotic linear behaviors, estimated with the last 10 experimental measurements, range from 0.10 for the RK4 method to 0.13 for the WEC one. Finally, an iteration takes respectively 4.26, 4.24 and 8.64 seconds for the proposed algorithm, the WEC variant and the RK4 one. Therefore the initially detailed algorithm in section 2.4 seems to achieve the best tradeoff between accuracy and speed in the considered experimental setting.

### 3.4 Prediction performance

Table 1 gathers the predictive performance of the proposed exp-parallelization method. The performance metric is the Dice coefficient, which ranges from 0 for disjoint structures to 1 for a perfect match. A Mann-Witney test is performed to quantify the significance of the results in comparison to a naive methodology, which keeps constant the baseline structures over time. Considering the very high dimension of the manifold, failing to accurately capture the disease progression

|  | **Predicted follow-up visit** | | | | | | |
|---|---|---|---|---|---|---|---|
| **Method** | M12 | M24 | M36 | M48 | M60 | M72 | M96 |
|  | N=140 | N=134 | N=123 | N=113 | N=81 | N=62 | N=17 |
| [exp] | .882 | .852 | **.825** $\}_*$ | **.796** $\}_{**}$ | **.768** $\}_*$ | **.756** $\}_{**}$ | **.730** $\}_*$ |
| [ref] | **.884** | .852 | .809 $\}^*$ | .764 $\}^{**}$ | .734 $\}^*$ | .706 $\}^{**}$ | .636 $\}^*$ |

Table 1: Averaged Dice performance measures. In each cell, the first line gives the average performance of the exp-parallelization-based prediction [exp], and the second line the reference one [ref]. Each column corresponds to an increasingly remote predicted visit from baseline. Significance levels [.05, .01, .001].

trend can quickly translates into unnatural predictions, much worse than the naive approach.

The proposed paradigm significantly outperforms the naive prediction three years or later from the baseline, thus demonstrating the relevance of the exp-parallelization concept for disease progression modeling, made computationally tractable thanks to the operational qualities of the fanning scheme for high-dimensional applications.

## 4 Conclusion

We detailed the fanning scheme for parallel transport on a high-dimensional manifold of diffeomorphisms, in the shape analysis context. Our analysis unveiled the operational qualities and computational efficiency of the scheme in high dimensions, with a empirical relative error below 1% for 25 steps only. We then took advantage of the parallel transport for accurately predicting the progression of brain structures in a personalized way, from previously acquired knowledge.